\pdfoutput=1

\documentclass[11pt]{article}

\usepackage[]{acl}

\usepackage{times}
\usepackage{latexsym}
\usepackage{stfloats}

\usepackage[T1]{fontenc}

\usepackage[utf8]{inputenc}
\usepackage{color}
\usepackage{csquotes}

\usepackage{microtype}

\title{Which one is more toxic? Findings from Jigsaw Rate Severity of Toxic Comments}

\author{Millon Madhur Das, Punyajoy Saha, Mithun Das \\
        Indian Institute of Technology, Kharagpur, India\\
        millonmadhurdas@kgpian.iitkgp.ac.in, \{punyajoys, mithundas\}iitkgp.ac.in}

\begin{document}
\maketitle
\begin{abstract}
The proliferation of online hate speech has necessitated the creation of algorithms which can detect toxicity. Most of the past research focuses on this detection as a classification task, but assigning an absolute toxicity label is often tricky. Hence, few of the past works transform the same task into a regression. This paper shows the comparative evaluation of different transformers and traditional machine learning models on a recently released toxicity severity measurement dataset by Jigsaw. We further demonstrate the issues with the model predictions using explainability analysis.

\textbf{Note:} \textit{This paper contains examples of toxic posts. But owing to the nature of work, we cannot avoid them.} 
\end{abstract}

\section{Introduction}

In social media, toxic language denotes a text containing inappropriate language in a post or a comment. The presence of toxic language on social media hampers the fabric of communication in the social media posts; e.g., toxic posts targeting some community might silence members of the community.  Subsequently, social media platforms like Facebook~\cite{Facebook43:online} and Twitter~\cite{Twitters6:online} have laid down moderation guidelines. They also employ various automatic and manual detection techniques to detect such forms of language and apply appropriate moderation~\cite{Updateon68:online}. Henceforth, researchers have started looking into this direction. Most of the past research focused on developing a classification task which again varies based on the classification labels the researchers choose, i.e., abusive or no abusive~\cite{nobata2016abusive}, hate speech, offensive and normal~\cite{mathew2021hatexplain}. This variation in the classification labels makes transferring models across different datasets tricky. Secondly, assigning a label to a post in terms of toxicity labels is complicated as many of the posts can be subjective~\cite{aroyo2019crowdsourcing}. Finally, a further challenge is that after encountering several highly toxic comments, an annotator might find subsequent moderately toxic comments as not toxic~\cite{kurrek-etal-2020-towards}. 

Research is currently trying to situate the toxicity detection tasks as regression tasks. In its simplest form, an annotator is provided two samples, and they have to decide which one is more toxic. Eventually, these annotated comparisons are converted to a scalar value which denotes the level of the toxicity of the post. ~\citet{hada-etal-2021-ruddit} uses best-worst scaling~\cite{kiritchenko-mohammad-2017-best} to assign toxicity scores to a post based on the comparison annotated by annotators. Besides, another study~\cite{kennedy2020constructing} used Rasch measurement theory for converting the comparisons to scalar values.

In this shared task, Jigsaw released a new dataset for understanding the severity of toxic language. The organizers select a set of 14,000 datapoints. They used these datapoints to create multiple pairs, which were then annotated by some annotator. The annotators marked one of the comments as toxic based on their notion of toxicity. These comparisons were compared with the ones received from models, and average agreement was used as the final score.

In this paper, we focus on developing models for this task. Since the shared task did not provide any training dataset, we utilized different classification-based toxic language datasets and converted their labels to a scalar value based on various strategies. Finally, we use simple models like TF-IDF to complex models like Transformers. We conclude the paper with a detailed error analysis to understand the behavior of the models.

\section{Datasets}

In this section, we illustrate the datasets used for this task. The first section~\ref{sec:task} describes the task dataset, and the second section~\ref{sec:ext} exhibits the dataset used for training the models since we don't have any training dataset associated with this task.

\subsection{Task dataset}
\label{sec:task}

In the task dataset~\footnote{https://www.kaggle.com/c/jigsaw-toxic-severity-rating}, pairs of comments were presented to expert raters, who marked one of two comments more harmful – each according to their notion of toxicity. The final label for each pair is decided with a majority vote. The validation dataset contains  $\sim$ 30k data points where each datapoint was a pair of toxic posts with the annotation mentioning which one is more toxic. Apart from this we were provided with 5\% of the test dataset for validating our models. The rest, 95\%, is private and was used as hidden test data. Our results are discussed for the validation dataset and entire test dataset (150k posts).

\subsection{External datasets}
\label{sec:ext}

\subsubsection{Ruddit}

This dataset~\cite{hada-etal-2021-ruddit} contains English language Reddit comments that have fine-grained, real-valued scores between -1 (maximally supportive) and 1 (maximally offensive). The annotators were given a set of 4 comments and asked to arrange them in order of their toxicity/abusiveness. These were converted to scalar scores using best-worst scaling~\cite{kiritchenko-mohammad-2017-best}. We transformed these scores to a value between 0 and 1 to keep the distribution of values uniform to other datasets. This dataset contains  $\sim$ 16k data points.

\subsubsection{Jigsaw Toxic Comment Dataset(JTC)}

This dataset contains a large number of Wikipedia comments labeled by human raters for toxic behavior. The types of toxicity are toxic, severe toxic, obscene, threat, insult, and identity hate. Each comment can have any one or more of these labels. It contains $\sim$230k data points. This dataset is a part of the Toxic Comment Classification Challenge hosted on Kaggle~\footnote{\url{https://tinyurl.com/2p85bsnj}}. We converted the labels into a single score. The different toxicity categories were given different weights, and the final toxicity score was the sum of weights for each example.
Our final weighing scheme was,
severe toxic:12, identity hate:9, threat:8, insult:6, obscene:5, toxic:4


\subsubsection{Jigsaw Unintended Bias Dataset}
This dataset is part of a Kaggle Competition, Jigsaw Unintended Bias in Toxicity Classification~\footnote{\url{https://tinyurl.com/9cbyp3ry}}. Each comment has a toxicity label that lies between 0 and 1. It has  $\sim$ 2 million samples. This attribute (and all others) are fractional values representing the fraction of human raters who believed the attribute applied to the given comment. For evaluation, test set examples with a target >= 0.5 will be considered to be in a positive class (toxic).

The data also has several toxicity sub-type attributes like severe toxicity, obscene, threat, insult, identity attack, and sexually explicit. We have used mapping similar to that used for the Jigsaw Toxic Comment dataset for assigning the toxicity score.

\subsubsection{Davidson}
The dataset is sourced from ~\cite{davidson2017automated}.
The data is compiled using a hate speech lexicon, and all the instances are from Twitter. A minimum of 3 coders labeled tweets into classes Hate speech, Offensive, and Neither. The final sample consisted of  $\sim$ 24,000 examples, and only about 5\% fell into the Hate Speech class. 
We map the toxicity score using the formula - (3$*$(\# hate speech annotations)+2$*$(\# offensive annotations)+(\# neither annotations))$/$No.of labelers. We then normalise this value between 0 and 1.

\subsubsection{Founta}
Similar to the previous dataset, ~\cite{founta2018large} analyzed comments from Twitter and published a dataset with  $\sim$ 80k examples. It has three labels (0, 1, 2) with an increasing level of toxicity. We scaled it between 0 and 1 by normalizing it. 

\section{Methodology}
We preprocessed the datasets using standard techniques like stemming, lemmatization, removing contractions, and hyperlinks. For the toxic severity rating, we first tried traditional techniques like TF-IDF~\cite{rajaraman_ullman_2011} and doc2vec~\cite{le2014distributed} based regressors to set the baseline. We further add other deep learning setups based on Transformers~\cite{vaswani2017attention} to check if the scores improve further.

\subsection{Baselines}

Initially, we used TF-IDF and Doc2Vec as feature extractors. \textbf{TFIDF} is a method to find the importance of a word to a document in a text corpus~\cite{rajaraman_ullman_2011}. Doc2Vec is an unsupervised method to represent a document as a vector. To train using these features, we use ridge regression, which enhances linear regression by adding L2 regularization. 

We used a hyperparameter optimization framework, Optuna, to automate the hyperparameter search for TFIDF. We found the Tfidf vectorizer to work best with the `charwb' analyzer, n-gram range (3,5) \&  vocabulary of  $\sim$ 30k most frequent words. The ridge regressor had a regularization strength of  $\sim$ 1. 

Doc2Vec was trained with a feature vector of size 300, learning rate $\alpha$ of 0.025. Both distributed memory and distributed bag of words methods were tested. As the performance was unsatisfactory, we did not conduct hyperparameter tuning for doc2vec.

\subsection{Transformers}

We take a pre-trained transformers model~\cite{vaswani2017attention} that outputs a 768-dimensional vector representation of an input sentence. As this output cannot be directly used as a score for toxicity, we added a single linear layer on top of the encoder to get a single value for toxicity. As we feed input data, the entire pre-trained transformers model and the additional untrained regression layer is trained on our specific task. We focused on tuning hyperparameters manually instead of using any hyperparameter search library due to resource constraints. All the transformers were trained for three epochs with a batch size of 16.

In the following section, we discuss the specifics of the pre-trained models used in detail.

\subsubsection{\href{https://huggingface.co/bert-base-multilingual-cased}{bert-base-multilingual-cased (M-BERT)}}
This language representation model is a modification of BERT, introduced by ~\cite{devlin2018bert}. It was pretrained on a large corpus of multilingual data from Wikipedia with the objective of Masked language modeling(MLM) in a self-supervised setting. In the masked language model pre-training, the model learns using predicting some of the mask tokens in the text, and it should also be noted that this model is case sensitive.

\subsubsection{\href{https://huggingface.co/bert-base-uncased}{bert-base-uncased (BERT)}}
Similar to the above model, this was also pretrained using MLM objective, except this model was trained only on English text corpus, specifically on the BookCorpus, and is not case sensitive.

\subsubsection{\href{https://huggingface.co/Hate-speech-CNERG/dehatebert-mono-english}{Hate-speech-CNERG/dehatebert-mono-English(dehateBERT)}}
~\cite{aluru2020deep} benchmarked hate speech classification models for 9 different languages and 16 datasets. All their models are based on the multilingual BERT model. We used their model that was finetuned on an English text corpus. 

\subsubsection{\href{https://huggingface.co/cardiffnlp/twitter-roberta-base-hate}{cardiffnlp/twitter-roberta-base-hate(HRoBERTa)}}
This model is derived from the roBERTa-base model~\cite{liu2019roberta} trained on $\sim$ 58M tweets and finetuned on for hate speech detection with the TweetEval benchmark ~\cite{barbieri2020tweeteval}. Unlike the previous two models, this is an end-to-end regression model, meaning given a sentence, it directly outputs a number between 0 and 1.

\subsubsection{\href{https://huggingface.co/GroNLP/hateBERT}{GroNLP/hateBERT}}
This is a re-trained BERT model for abusive language detection in English by ~\cite{caselli2020hatebert}. It was trained using MLM objective on RAL-E, a large-scale dataset of Reddit comments in English.

\subsubsection{\href{https://huggingface.co/sentence-transformers/all-mpnet-base-v2}{sentence-transformer/mpnet-base-v2(mpnet)}}
This is a sentence embedding model introduced by ~\cite{reimers2019sentence} trained using a self-supervised contrastive learning objective. It is trained on 1 billion sentence pairs and is based on the pretrained Masked and Permuted Network introduced by ~\cite{song2020mpnet}. It solves the problems of MLM in BERT and PLM (permuted language modeling) in XLNet and achieves better accuracy.



\subsection{Ensembles}
Finally, we experimented with ensembles of the models described in the previous sections. To do the ensembling, we predict the scores for a typical post using various models and then combine the scores using a weighted average. The weights are decided based on the performance of the validation dataset. We used the weights as a variable using the Limited-memory BFGS (LM-BFGS) method, which is an optimization function in the family of quasi-Newton methods that approximates the Broyden–Fletcher–Goldfarb–Shanno algorithm (BFGS) using a limited amount of computer memory. It is a popular algorithm for parameter estimation in machine learning. The algorithm's target problem is to minimize $f(x)$ over unconstrained values of the real-vector $x$  where $f$ is a differentiable scalar function.

\section{Results and Inference}
In this section, we present a detailed analysis of the performance of our models.

\subsection{Comparative study of performance}
Table~\ref{tab:performance} shows the performance of our model on the validation dataset and total test dataset. As expected, the transformer-based approaches outperform the traditional approaches like TF-IDF/doc2vec. We found that HRoBERTa model performed the best among the transformers models. It is interesting to note that BERT \& M-BERT give comparable results to language models already pretrained for detecting toxicity(hateBERT \& dehateBERT). Experiments on the transformed Founta, Davidson, and Toxic unintended did not give good scores; hence we did not perform further experiments on them.

Our team secured a rank of 145 out of 2301 in the Kaggle Jigsaw Rate Severity of Toxic Comments Competition with an accuracy of 79.84\% in the private leaderboard. However, one of our ensembles which was not part of our final submission, performed even better. We achieved an accuracy of 80.74\% in the final standings (Table~\ref{tab:performance}). It is also worth mentioning that our approach was quite similar to the winning approach(accuracy of 81.39\%), except they used Genetic Algorithm~\cite{GuanshuoXu} to find weights for their ensemble. Our method using an ensemble of 5 models performs half a percent worse than their 15 ensemble model.

\begin{table*}
\centering
\scriptsize
\begin{tabular}{ll}

\textbf{Less Toxic Sentence} & \textbf{More Toxic Sentence} \\
\hline
\hline

{\setlength{\fboxsep}{0pt}\colorbox{white!0}{\parbox{0.45\textwidth}{
\colorbox{red!21.736497789910633}{\strut real} \colorbox{red!25.771625110386694}{\strut encyclopedia} \colorbox{red!0.0}{\strut So} \colorbox{red!100.00000000000001}{\strut f**k} \colorbox{red!19.42056208123472}{\strut you} \colorbox{red!79.63365614362758}{\strut f**king} \colorbox{red!0.0}{\strut all} \colorbox{red!50.805432617382706}{\strut f*g} \colorbox{red!93.94713708891688}{\strut f**ks} 
}}} 

& 

{\setlength{\fboxsep}{0pt}\colorbox{white!0}{\parbox{0.45\textwidth}{
\colorbox{red!100.00000000000001}{\strut vandalism} \colorbox{red!0.0}{\strut will} \colorbox{red!48.2811904852199}{\strut you} \colorbox{red!0.0}{\strut please} \colorbox{red!0.0}{\strut stop} \colorbox{red!0.0}{\strut vandalizing} \colorbox{red!0.0}{\strut my} \colorbox{red!0.0}{\strut talk} \colorbox{red!0.0}{\strut page} \colorbox{red!0.0}{\strut ?} \colorbox{red!0.0}{\strut it} \colorbox{red!0.0}{\strut very} \colorbox{red!31.916409734506583}{\strut irritating} \colorbox{red!26.13337735930517}{\strut and} \colorbox{red!0.0}{\strut borderlining} \colorbox{red!35.34008909452624}{\strut outright} \colorbox{red!0.0}{\strut harassment} 
}}} \\

\hline

{\setlength{\fboxsep}{0pt}\colorbox{white!0}{\parbox{0.45\textwidth}{
\colorbox{red!0.0}{\strut Hey} \colorbox{red!100.0}{\strut f*ck} \colorbox{red!36.91739956233456}{\strut you} \colorbox{red!89.14944737394667}{\strut c*nt} \colorbox{red!0.0}{\strut bag} \colorbox{red!0.0}{\strut !} \colorbox{red!36.91739956233456}{\strut you} \colorbox{red!0.0}{\strut dont} \colorbox{red!0.0}{\strut own} \colorbox{red!0.0}{\strut wikpedia} \colorbox{red!0.0}{\strut and} \colorbox{red!27.10192540435746}{\strut please} \colorbox{red!0.0}{\strut report} \colorbox{red!64.60628574376894}{\strut my} \colorbox{red!0.0}{\strut i} \colorbox{red!0.0}{\strut want} \colorbox{red!0.0}{\strut to} \colorbox{red!0.0}{\strut close} \colorbox{red!64.60628574376894}{\strut my} \colorbox{red!0.0}{\strut account} \colorbox{red!0.0}{\strut any} \colorbox{red!0.0}{\strut way} \colorbox{red!36.91739956233456}{\strut you} \colorbox{red!27.36771790531552}{\strut creep} 
}}} 

& 

{\setlength{\fboxsep}{0pt}\colorbox{white!0}{\parbox{0.45\textwidth}{
\colorbox{red!0.0}{\strut Please} \colorbox{red!53.59493627598432}{\strut stop} \colorbox{red!0.0}{\strut your} \colorbox{red!54.954221877574355}{\strut disruptive} \colorbox{red!0.0}{\strut editing} \colorbox{red!0.0}{\strut If} \colorbox{red!0.0}{\strut your} \colorbox{red!0.0}{\strut vandalism} \colorbox{red!0.0}{\strut continues} \colorbox{red!0.0}{\strut you} \colorbox{red!0.0}{\strut will} \colorbox{red!86.1636021538142}{\strut be} \colorbox{red!0.0}{\strut blocked} \colorbox{red!0.0}{\strut from} \colorbox{red!0.0}{\strut editing} \colorbox{red!0.0}{\strut Wikipedia} \colorbox{red!100.0}{\strut andemu} 
}}}\\

\hline

{\setlength{\fboxsep}{0pt}\colorbox{white!0}{\parbox{0.45\textwidth}{
\colorbox{red!39.55778383443449}{\strut i} \colorbox{red!0.0}{\strut said} \colorbox{red!82.32015736605149}{\strut P**S} \colorbox{red!78.33766715149144}{\strut OFF} \colorbox{red!16.901106475672698}{\strut YOU} \colorbox{red!32.06496706586953}{\strut GEEKY} \colorbox{red!42.860007337618505}{\strut LITTLE} \colorbox{red!100.0}{\strut S**T} 
}}}

& 

{\setlength{\fboxsep}{0pt}\colorbox{white!0}{\parbox{0.45\textwidth}{
\colorbox{red!50.56006721675743}{\strut And} \colorbox{red!0.0}{\strut yes} \colorbox{red!0.0}{\strut people} \colorbox{red!0.0}{\strut should} \colorbox{red!0.0}{\strut recognize} \colorbox{red!0.0}{\strut that} \colorbox{red!0.0}{\strut but} \colorbox{red!0.0}{\strut they} \colorbox{red!0.0}{\strut usually} \colorbox{red!0.0}{\strut do} \colorbox{red!0.0}{\strut not} \colorbox{red!0.0}{\strut One} \colorbox{red!0.0}{\strut of} \colorbox{red!0.0}{\strut the} \colorbox{red!0.0}{\strut first} \colorbox{red!47.96182006764966}{\strut objections} \colorbox{red!33.99243529016303}{\strut you} \colorbox{red!0.0}{\strut hear} \colorbox{red!0.0}{\strut directed} \colorbox{red!0.0}{\strut at} \colorbox{red!0.0}{\strut someone} \colorbox{red!0.0}{\strut who} \colorbox{red!0.0}{\strut says} \colorbox{red!0.0}{\strut he} \colorbox{red!0.0}{\strut favors} \colorbox{red!0.0}{\strut free} \colorbox{red!0.0}{\strut markets} \colorbox{red!0.0}{\strut is} \colorbox{red!0.0}{\strut invariably} \colorbox{red!0.0}{\strut There} \colorbox{red!0.0}{\strut no} \colorbox{red!0.0}{\strut such} \colorbox{red!0.0}{\strut thing} \colorbox{red!0.0}{\strut as} \colorbox{red!0.0}{\strut a} \colorbox{red!0.0}{\strut free} \colorbox{red!0.0}{\strut market} \colorbox{red!100.0}{\strut Ridiculously} \colorbox{red!0.0}{\strut trivial} 
}}}\\

\hline
\hline

{\setlength{\fboxsep}{0pt}\colorbox{white!0}{\parbox{0.45\textwidth}{
\colorbox{red!100.0}{\strut F**k} \colorbox{red!0.0}{\strut up.}
}}}

&

{\setlength{\fboxsep}{0pt}\colorbox{white!0}{\parbox{0.45\textwidth}{
\colorbox{red!0.0}{\strut Wikinfo,} \colorbox{red!0.0}{\strut and} \colorbox{red!0.0}{\strut purge} \colorbox{red!0.0}{\strut devils} \colorbox{red!0.0}{\strut from} \colorbox{red!0.0}{\strut here.} \colorbox{red!0.0}{\strut Muzemike} \colorbox{red!100.0}{\strut damned} \colorbox{red!0.0}{\strut for} \colorbox{red!0.0}{\strut being} \colorbox{red!0.0}{\strut enemy} \colorbox{red!0.0}{\strut of} \colorbox{red!0.0}{\strut God:} \colorbox{red!0.0}{\strut I} \colorbox{red!0.0}{\strut was} \colorbox{red!0.0}{\strut banned} \colorbox{red!0.0}{\strut for} \colorbox{red!0.0}{\strut this:} 
\colorbox{red!0.0}{\strut So} \colorbox{red!0.0}{\strut be} \colorbox{red!0.0}{\strut damned,} \colorbox{red!0.0}{\strut MuZemike,} \colorbox{red!0.0}{\strut daemon} \colorbox{red!0.0}{\strut in} \colorbox{red!66.8229162373972}{\strut human} \colorbox{red!0.0}{\strut skin.} 
}}}\\

\hline

{\setlength{\fboxsep}{0pt}\colorbox{white!0}{\parbox{0.45\textwidth}{
\colorbox{red!0.0}{\strut Why} \colorbox{red!0.0}{\strut don't} \colorbox{red!7.64260920393303}{\strut you} \colorbox{red!9.14030969532675}{\strut go} \colorbox{red!100.0}{\strut s*rew} \colorbox{red!0.0}{\strut Why} \colorbox{red!0.0}{\strut don't} \colorbox{red!7.64260920393303}{\strut you} \colorbox{red!9.14030969532675}{\strut go} \colorbox{red!100.0}{\strut s*rew} \colorbox{red!0.72032005424685}{\strut instead} \colorbox{red!0.0}{\strut of} \colorbox{red!12.914743230731183}{\strut harassing} \colorbox{red!0.0}{\strut me?} 
}}}

&

{\setlength{\fboxsep}{0pt}\colorbox{white!0}{\parbox{0.45\textwidth}{
\colorbox{red!0.0}{\strut I} \colorbox{red!0.0}{\strut also} \colorbox{red!0.0}{\strut think} \colorbox{red!0.0}{\strut ...}  \colorbox{red!71.36842784689311}{\strut For} \colorbox{red!0.0}{\strut example} \colorbox{red!0.0}{\strut i} \colorbox{red!0.0}{\strut can} \colorbox{red!0.0}{\strut say} \colorbox{red!0.0}{\strut that} \colorbox{red!100.0}{\strut Muslims} \colorbox{red!0.0}{\strut go} \colorbox{red!0.0}{\strut to} \colorbox{red!0.0}{\strut mosques,} \colorbox{red!0.0}{\strut pray,} \colorbox{red!0.0}{\strut beat,} \colorbox{red!0.0}{\strut up} \colorbox{red!0.0}{\strut their} \colorbox{red!0.0}{\strut wives,} \colorbox{red!0.0}{\strut blow,} \colorbox{red!0.0}{\strut up,} \colorbox{red!0.0}{\strut etc,} \colorbox{red!0.0}{\strut what} \colorbox{red!0.0}{\strut do} \colorbox{red!0.0}{\strut atheists} \colorbox{red!0.0}{\strut do?.}  
}}}

\end{tabular}
\caption{\scriptsize{Samples mislabeled by human labeler (top 3) and model misclassifications (bottom 2). The highlighted text denotes how words affect the model predictions. Darker highlight denotes that the model is paying more attention to that words.}}
\label{tab:examples}
\end{table*}

\begin{table}[!htpb]
\scriptsize
\centering
\begin{tabular}{llll}
\textbf{Dataset} & \textbf{Models} & \textbf{Val. Acc.} & \textbf{Test Acc.}\\
\hline
& TF-IDF & 57.54 & 69.38\\
& M-BERT & 59.83 & 74.71 \\
Ruddit & BERT & 60.71 & 78.41 \\ 
& HRoBERTa (A) & \textbf{61.06} & \textbf{79.47} \\ 
& hateBERT & 60.69 & 78.46 \\
& dehateBERT & 58.52 & 71.28 \\ 

\hline

& TF-IDF & 61.01 &  78.57\\ 
& doc2vec &  59.87 & 68.80 \\
& M-BERT (B) & 61.31 & 79.17 \\
JTC & BERT (C) & 61.32 & 78.79 \\ 
& HRoBERTa (D) & \textbf{61.53} & \textbf{80.16} \\ 
& hateBERT (E) & 61.25 & 78.90 \\
& dehateBERT & 59.81 & 74.95 \\ 
\hline

Founta & TF-IDF & 64.58 & 72.66\\ 
 & BERT & 51.50 & 75.67 \\

\hline
Toxic & TF-IDF & 62.64 &  72.47\\ 
Unintended& BERT & 59.92 & 77.70\\
\hline
Davidson & TF-IDF & 62.64 &  72.47\\ 
& BERT & 52.38 & 76.64\\

\hline
\multicolumn{2}{c}{\textbf{A+B+C+D+E}} & \textbf{76} & \textbf{80.74} \\\hline

\end{tabular}
\caption{Performance on Jigsaw Rate Severity of Toxic Comment Dataset for the validation and entire test dataset.}
\label{tab:performance}
\end{table}

\subsection{LIME}
We also conducted local interpretable model-agnostic explanations extensively on our best model (HRoBERTa) to identify potential issues with model predictions on the validation dataset. The validation set contains pairs of sentences labeled as less toxic and more toxic.

We ranked the model predictions and checked the top 100 wrong predictions manually. The top 100 wrong predictions were found by ranking the difference between the score assigned to less toxic to more toxic sentence. For most of the cases, it was not the model but the human annotator who was at fault. There were several cases where we found difficult to select the more toxic comment. We found 68 samples where the annotator was wrong, 3 samples where our model was wrong and found 29 samples to be equally toxic. We add some of the samples from each category in Table~\ref{tab:examples}. 

\bibliography{anthology,custom}
\bibliographystyle{acl_natbib}

\appendix

\end{document}